\documentclass{bmvc2k}

\title{Efficient Vision Transformer for Human Pose Estimation via Patch Selection}

\addauthor{Kaleab A. Kinfu}{kinfu@seas.upenn.edu}{1}
\addauthor{Ren\'e Vidal}{vidalr@seas.upenn.edu}{1}

\addinstitution{
 Center for Innovation in Data Engineering and Science (IDEAS),
 University of Pennsylvania\\
 Philadelphia, PA, USA
}

\runninghead{Kinfu, Vidal}{Efficient Vision Transformer for Human Pose Estimation}

\def\etal{\emph{et al}\bmvaOneDot}
\def\ie{\emph{i.e}\bmvaOneDot}

\usepackage{amsmath}
\usepackage{amssymb}

\usepackage{algpseudocode}
\usepackage{algorithm}
\usepackage{breqn}

\usepackage{epsfig}
\usepackage{graphicx}
\usepackage{booktabs}
\usepackage{enumitem}
\usepackage{caption}
\usepackage{subcaption}
\usepackage{parskip}

\newcommand{\myparagraph}[1]{\smallskip\noindent\textbf{#1.}}

\begin{document}
\def\method{{EViTPose}}
\maketitle

\begin{abstract}
   While Convolutional Neural Networks (CNNs) have been widely successful in 2D human pose estimation, Vision Transformers (ViTs) have emerged as a promising alternative to CNNs, boosting state-of-the-art performance. However, the quadratic computational complexity of ViTs has limited their applicability for processing high-resolution images. In this paper, we propose three methods for reducing ViT's computational complexity, which are based on selecting and processing a small number of most informative patches while disregarding others. The first two methods leverage a lightweight pose estimation network to guide the patch selection process, while the third method utilizes a set of learnable joint tokens to ensure that the selected patches contain the most important information about body joints. Experiments across six benchmarks show that our proposed methods achieve a significant reduction in computational complexity, ranging from 30\% to 44\%, with only a minimal drop in accuracy between 0\% and 3.5\%. 
\end{abstract}

\section{Introduction}
\label{sec:intro}
In recent years, Human Pose Estimation has emerged as an important problem in computer vision, with numerous applications in fields such as surveillance~\cite{Lamas2022HumanPE}, motion analysis~\cite{Stenum2020TwodimensionalVA}, virtual and augmented reality~\cite{Obdrzlek2012RealTimeHP, Lin2010AugmentedRW}. Classical pose estimation algorithms relied on handcrafted features
\cite{Gkioxari2013ArticulatedPE, Ramanan2006LearningTP, Johnson2010ClusteredPA}, but recent advances in deep learning have led to significant improvements based on learned features~\cite{Toshev2014DeepPoseHP, Yu2021LiteHRNetAL}. For instance, Convolutional Neural Networks (CNNs) have proven to be successful by exploiting spatial correlations among pixels. 

The recent emergence of Vision Transformers (ViTs) has challenged the dominance of CNNs. Unlike CNNs, ViTs rely on self-attention mechanisms to model the long-range dependencies between patches, which has been shown to be highly effective~\cite{Dosovitskiy2021AnII}. Nevertheless, the computational complexity of ViTs presents a significant challenge for processing high-resolution images. In particular, the computational cost of ViTs scales quadratically with the number of input tokens, making them intractable for practical use.

To address this issue, several recent works have proposed various methods for reducing the number of tokens that need to be processed by ViTs, thereby lowering their computational cost. Token Learner~\cite{Ryoo2021TokenLearnerWC} is one approach that aims to merge and reduce the input tokens into a small set of important learned tokens. Token Pooling~\cite{Marin2021TokenPI} clusters the tokens and down-samples them, whereas DynamicViT~\cite{Rao2021DynamicViTEV} introduces a token scoring network to identify and remove redundant tokens. Although these techniques successfully reduce the computational complexity of ViTs in classification tasks, the additional pooling and scoring network can introduce extra computational overhead. Besides, the extension of these approaches to dense prediction tasks, such as human pose estimation, remains an open question.

In this work, we propose to reduce the computational complexity of transformer based human pose estimation networks by selecting and processing a small subset of patches that are most likely to contain body joints or limbs. The rationale behind this choice is that results from~\cite{Yang2021TransPoseKL} suggest that pixels neighboring a joint or a limb 
are more informative about joint locations.
We first propose two patch selection methods that utilize fast yet imprecise out-of-the-shelf pose estimation networks to guide the selection process. We then propose a patch selection method that uses learnable joint tokens to progressively select the most informative patches. As a result, our approach significantly enhances computational efficiency, 
albeit with a minor trade-off in accuracy. To validate the effectiveness of our proposed methods, we conducted experiments on six 2D human pose estimation benchmarks.
The experiments show that our approach significantly reduces computational complexity (30\% to 44\% drop in GFLOPs), while only having a minimal drop in accuracy (between 0\% and 3.5\%).


\begin{figure}[t]
\centering
  \begin{subfigure}[b]{0.7\textwidth}
\includegraphics[width=1.\textwidth]{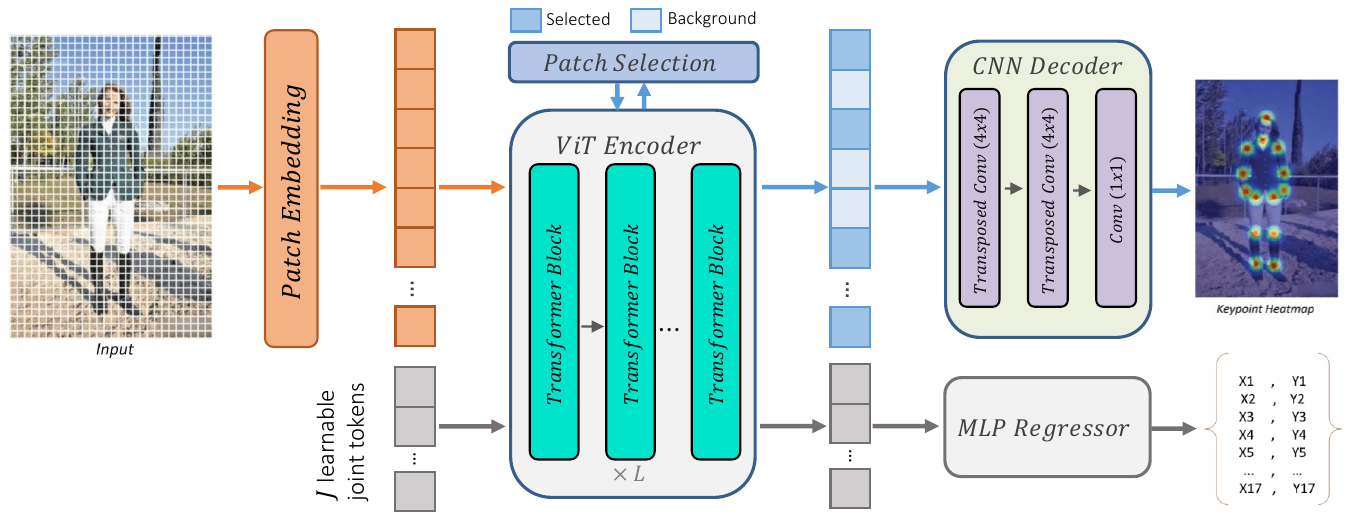}
    \caption{}
    \label{fig:vitpose}
  \end{subfigure}
   \hspace{0.001\textwidth}
  \color{gray}
  \rule[3.4ex]{0.05pt}{22ex} 
  \color{black}
  \hspace{0.01\textwidth}
  \begin{subfigure}[b]{0.125\textwidth}
    \includegraphics[width=1.0\textwidth]{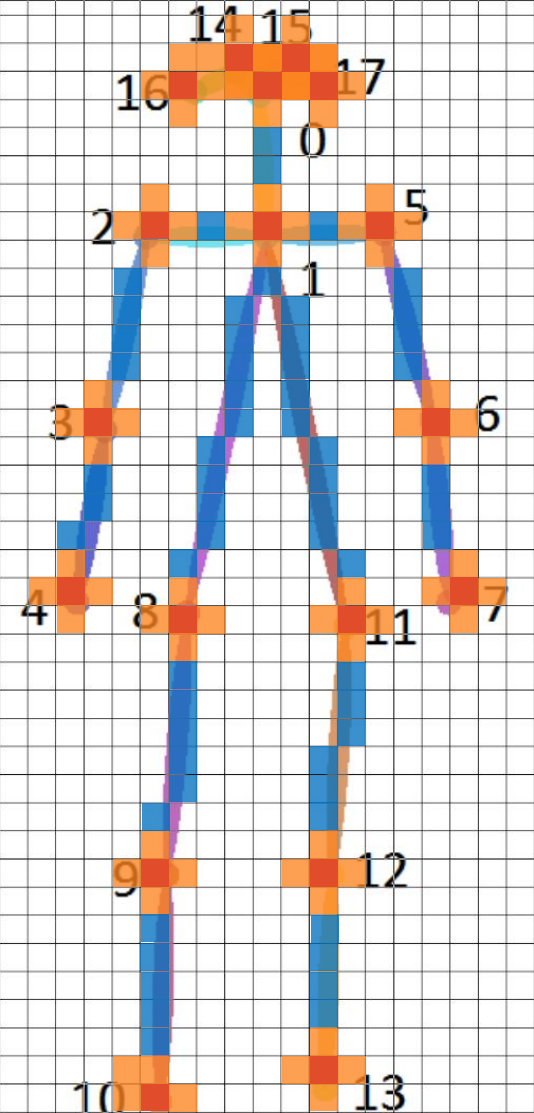}
    \caption{}
    \label{fig:skel}
  \end{subfigure}
  \caption{\textbf{Overall architecture of
  \method\
  -- } (a) The input image is first passed through a patch embedding layer to obtain patches of size $16\times 16$. These patches, along with $J$ learnable joint tokens, are processed by a ViT with $L$ transformer blocks. Patches are selected either before processing with ViT for neighboring and skeleton techniques or progressively across blocks for joint tokens based patch selection. The output of the last ViT block is then used by a CNN-based decoder to estimate the heatmap of $J$ joints, while a simple MLP joint regressor estimates joints directly from joint tokens. (b) Skeletal visualization of selected patches via the proposed patch selection methods. The red, orange, and blue patches correspond to the 
body joints, neighboring patches, and skeleton patches, respectively.}
  \label{fig:vitpose_skel}
  
\end{figure}

\section{Related Work}
\label{sec:related}


\subsection{Human Pose Estimation}
Human Pose Estimation is an essential task in computer vision that involves identifying and estimating the location of human body joints from 2D images or videos. 
In recent years, deep learning methods have been successful in 2D human pose estimation, with most methods employing CNNs to learn a mapping between the input image and the corresponding 2D pose. 
There are two main deep learning pipelines used in single person pose estimation: regression-based and heatmap-based approaches. Regression-based methods directly map the input image to 2D joint positions~\cite{Toshev2014DeepPoseHP}, while heatmap-based methods predict approximate joint locations using 2D Gaussian heatmaps centered at the body joint~\cite{Wei2016ConvolutionalPM}. The heatmap-based approach is effective and commonly used, and will be used for inference in this work although both will be utilized for training.

Multi-person pose estimation is a difficult task that requires determining the number of people and their positions, as well as grouping key points. Two main approaches are top-down and bottom-up. Top-down approaches~\cite{Sun2019DeepHR, Xiao2018SimpleBF, Newell2016StackedHN, Yang2021TransPoseKL, Yuan2021HRFormerHT} use person detectors to extract boxes from input images and then apply single-person pose estimators to produce multi-person poses. In contrast, bottom-up techniques~\cite{Cao2021OpenPoseRM, Newell2017AssociativeEE, Cheng2020HigherHRNetSR, Shi2022EndtoEndMP} identify all body joints in a single image and group them for each person. This work will follow the top-down approach, which has been shown to be effective.

\subsection{Vision Transformer}
The Vision Transformer (ViT) architecture proposed by Dosovitskiy \etal~\cite{Dosovitskiy2021AnII} has demonstrated remarkable performance on image classification tasks. Consequently, several transformer based architectures have been proposed for human pose estimation. Transpose~\cite{Yang2021TransPoseKL} is a transformer network that estimates 2D pose using a CNN-based backbone.
TokenPose~\cite{Li2021TokenPoseLK} is another transformer based on explicit token representation for each body joint. HRFormer~\cite{Yuan2021HRFormerHT} is a transformer that adopts HRNet design along with convolution and local-window self-attention. ViTPose~\cite{Xu2022ViTPoseSV} is a ViT-based approach that uses a shared encoder training on multiple datasets to improve performance. These architectures demonstrate the effectiveness of transformer-based models in human pose estimation.

However, vision transformers have quadratic computational complexity. To improve the efficiency of ViT, researchers have proposed methods such as sparsifying the attention matrix~\cite{Roy2021EfficientCS, Child2019GeneratingLS}, token pooling~\cite{Marin2021TokenPI}, and estimating the significance of tokens~\cite{Rao2021DynamicViTEV}. Hierarchical Visual Transformer~\cite{Pan2021ScalableVT} removes redundant tokens via token pooling, while TokenLearner~\cite{Ryoo2021TokenLearnerWC} introduces a learnable tokenization module. 
Adaptive Token Sampler~\cite{Fayyaz2022AdaptiveTS} adaptively down-samples input tokens, which assigns significance scores to every token based on the attention weights of the class token in ViT. Similarly, EViT~\cite{Liang2022NotAP} determines tokens' importance scores via attention weights. However, it is still an open question whether these approaches can be extended to dense prediction tasks, such as human pose estimation. In our third patch selection method, we will follow a similar approach to \cite{Liang2022NotAP, Fayyaz2022AdaptiveTS} and extend it to the human pose estimation task.

\section{ViT based Human Pose Estimation with Patch Selection}
\label{sec:method}

In this section, we describe our approach to human pose estimation, which includes revisiting the standard ViT-based method, incorporating learnable joint tokens and  patch selection techniques. We present the overall architecture of our method in Figure~\ref{fig:vitpose}. 
Given an input image $X \in R^{H \times W \times 3}$, the task is to find a mapping from $X$ to $Y\in \mathbb{R}^{J\times 2}$, where $J$ is the number of body joints, for each person in the image. We first embed $X$ into patches of size $16\times 16$, resulting in a 
set of patch tokens $\mathbf{P}\in \mathbb{R}^{N\times C}$, 
where $N=\frac{H}{16} \times \frac{W}{16}$ and $C$ represents the channel dimension. We then extend the patch tokens with $J$ learnable joint tokens, $\mathbf{J} \in \mathbb{R}^{J\times C}$, that explicitly embed each one of the joints that are later used to regress the joint 2D positions in the image, resulting in a concatenated set of input tokens $\mathbf{I}\in \mathbb{R}^{(N+J)\times C}$. 
The patch and joint tokens are then fed to a standard ViT encoder with $L$ transformer blocks, each of which consists of a multi-head self-attention (MSA) layer and a feed-forward network (FFN). 

Here, we revisit the standard self-attention mechanism. Later, we will show how this mechanism is modified in the case of the joint-token-based patch selection method.
In the self-attention mechanism, the output tokens $\mathbf{O}$ and the attention matrix $\mathcal{A}$ are computed as
\begin{equation}
\mathbf{O} = \mathcal{A} \mathcal{V}
\quad\text{and}\quad
\mathcal{A} = \text{Softmax}( \frac{\mathcal{Q} \mathcal{K}^T}{\sqrt{C}}), 
\end{equation}
where $\mathcal{Q}\in \mathbb{R}^{(N+J)\times C}$, $\mathcal{K}\in \mathbb{R}^{(N+J)\times C}$ and $\mathcal{V} \in \mathbb{R}^{(N+J)\times C}$ are, respectively, the queries, keys, and values, which are computed from the input tokens $\mathbf{I}$ as in the standard ViT~\cite{Dosovitskiy2021AnII}.


Given the final feature map of the patches produced by ViT, denoted as $F_{out} \in \mathbb{R}^{ \frac{H}{16} \times \frac{W}{16} \times C}$, we use a classical decoder with two deconvolution blocks, each with a deconvolution layer, batch normalization, and ReLU activation to estimate the heatmap of $J$ joints. Similarly,  a LayerNorm layer followed by a fully connected MLP layer is used to directly regress the $J$ joint coordinates from the joint tokens. In this way, the joint tokens are enforced to learn the important joint-level information to be able to successfully regress the joint 2D positions.

\subsection{Improving Efficiency via Patch Selection}
Although ViT can model long-range dependencies and is able to generate a global representation of the overall image, the computational complexity increases quadratically with respect to the number of tokens. However, not all patches in an image contribute equally to the human pose estimation task. Recent research~\cite{Yang2021TransPoseKL} indicates that the long-range dependencies between predicted joints are mostly restricted to the body part regions. Therefore, computing MSA between every patch in the image while only a few patches are relevant to the body parts is unnecessary.
To this end, we propose three methods
to select a small number of relevant patches while discarding irrelevant and background patches without re-training the vision transformer. By selecting only the relevant patches, we can significantly reduce the computational complexity as shown in ~\cite{Rao2021DynamicViTEV, Fayyaz2022AdaptiveTS, Meng2021AdaViTAV} for the classification task.

The first two approaches utilize an off-the-shelf lightweight pose estimator to guide the patch selection. 
Our first approach is based on a breadth-first neighboring search algorithm that selects body joint patches and its neighbors given estimated pose predictions, as outlined in Section~\ref{sec:neighbor_patch_sel}. In our second approach, we extend the first approach to select patches formed by a skeleton of the joints. Here, the objective is to select body patches where the lines formed by body joint pairs cross. To accomplish this, we utilize Bresenham's algorithm to select the relevant patches, as outlined in Section~\ref{sec:skel_patch_sel}. It is important to note, however, that by selecting a few patches of the image and processing them with the ViT encoder, we only get the features of the patches that were chosen. However, we must create a feature map for all patches for further processing. As a result, since the goal of the task is to produce a Gaussian heatmap centered at the body joint and zero elsewhere, we fill the non-body-part patches with zeros.
Whilst the aforementioned methods can remove irrelevant patches before they are processed with ViT and thus enhance its efficiency, their reliance on the accuracy of off-the-shelf pose estimators is a limitation. As a result, we present an alternative approach for automatically selecting body part patches via learnable joint tokens that enable the selection of relevant patches using their corresponding attention maps, as outlined in Section~\ref{sec:auto_patch_sel}.

\begin{figure*}[t]
\begin{minipage}{0.48\textwidth}
\begin{algorithm}[H]
    \centering
    \caption{\bf (Select body joint patches and neighbors given joint prediction)}
\label{alg:Neighbor}
    \begin{algorithmic}[1]
\Require joint prediction $\mathcal{B} \in \mathbb{R}^{J\times 2}$, patch size $P$, neighboring search function $\mathcal{N}$ and $n$ number of neighboring patches.
	\Function{SelectJointPatches}{} 
	\State{ $BP$ = \{\}} \Comment{set of body part patches}
    \State{ $c = \frac{W}{P}$ } \Comment{number of columns}
    \For{$j \gets 1$ to $J$} 
        \State{ $x^j = \lfloor \frac{\mathcal{B}^j_x}{P} \rfloor$ } \Comment{get column}
        \State{$y^j= \lfloor \frac{\mathcal{B}^j_y}{P} \rfloor$  } \Comment{get row}
        
        \For{$ (k,l) \in \mathcal{N}(x^j, y^j, n) $} 
        \State{ $x^n = (x^j + k)$}
        \State{ $y^n = (y^j + l)$}
        
        \State {$ p = y^n \times c  + x^n$}
        \State {$BP \gets p$ } \Comment{add patch to set}
        \EndFor
        
    \EndFor
	\EndFunction
\end{algorithmic}
\end{algorithm}
\end{minipage}
\hfill
\begin{minipage}{0.48\textwidth}
\flushright
\begin{algorithm}[H]
    \caption{\bf (Select body part patches formed by a skeleton )}
    \label{alg:Bresenham}
    \begin{algorithmic}[1]
\Require joint prediction $\mathcal{B} \in \mathbb{R}^{J\times 2}$, body joint pairs $\mathcal{P}$, patch size $P$

	\Function{SelectSkeletonPatches}{} 
	\State{ $BP$ = \{\}} 
    \For{$j \gets 1$ to $J$} 
        
        \State{ $x_0^j = \lfloor \frac{\mathcal{B}^j_x}{P} \rfloor$,  $y_0^j= \lfloor \frac{\mathcal{B}^j_y}{P} \rfloor$} 
        \For{$ l \in \mathcal{P}(j)$} 
            \State{ $x_1^j = \lfloor \frac{\mathcal{B}^l_x}{P} \rfloor$, $y_1^j= \lfloor \frac{\mathcal{B}^l_y}{P} \rfloor$ } 

            \State{$\Delta_x = x_1 - x_0$, $\Delta_y = y_1 - y_0$}
            \State{$\epsilon = 2 \Delta_y - \Delta_x$} 
            \State{$y = y_0$}
            \For{$ x \in [x_0, x_1]$}
                \State {$BP \gets (x,y)$ } 
                \If{$\epsilon \geq 0$}
                    \State{$y = y + 1$}
                    \State{$\epsilon = \epsilon - 2\Delta_x $}
                \EndIf
            \State{$\epsilon = \epsilon + 2 \Delta_y$}
            \EndFor
        \EndFor
                
    \EndFor
	\EndFunction
\end{algorithmic}
\end{algorithm}
\end{minipage}
\end{figure*}

\subsection{Neighboring Patch Selection Method }
\label{sec:neighbor_patch_sel}
In our first approach, we assume we are provided with an estimate of the joint locations $\mathcal{B} \in \mathbb{R}^{J \times 2}$ via a lightweight pose estimation network, where $J$ denotes the number of joints. To select the body joint patch and its $n$ neighboring patches, we employ a Breadth-First Search (BFS) algorithm, as presented in Algorithm \ref{alg:Neighbor}. Specifically, for every joint located at 2D patch location $(x,y) \in \mathcal{B}$, we identify the nearest four neighboring patches located at $(x,y+1)$, $(x,y-1)$, $(x-1,y)$, and $(x+1,y)$, and store them in a queue. We continue searching for neighboring patches until we have selected $n$ patches, ensuring that we do not revisit any previously visited patches. To provide a visual representation of this approach, we depict in Figure~\ref{fig:skel} a skeletonized human figure where the red patches represent the joint patches and the set of orange patches correspond to the selected neighboring patches.

\subsection{Skeleton Patch Selection Method}
\label{sec:skel_patch_sel}
A limitation of the aforementioned neighboring selection method is that it only covers the body joint patches and its neighbors. Our second approach extends this method to encompass all patches between joints. This can lead to improved performance with a minimal increment in the computational complexity. To achieve this, we adopt a different strategy by 
identifying the patches where the line formed by the body joint pair (segments) crosses, based on Bresenham's algorithm~\cite{Bresenham1965AlgorithmFC}.
Originally proposed as a canonical line-drawing algorithm for pixellated grids, our extension of Bresenham determines the patches that need to be selected in the line between $(x_0, y_0)$ and $(x_1, y_1)$, corresponding to the start and end 2D locations of patches containing body joints. As we move across the $x-$ or $y-$ axis in unit intervals, we select the $x$ or $y$ value between the current and next value that is closer to the line formed by the body joint pairs. To make this decision, we require a parameter $\epsilon$. Our objective is to track the slope error from the last increment, and if the error exceeds a certain threshold, we increment our coordinate values and subtract from the error to re-adjust it to represent the distance from the top of the new patch, as presented in Algorithm \ref{alg:Bresenham}. 
As depicted in Figure~\ref{fig:skel}, the set of blue patches represents the patches selected by this approach.

\subsection{Joint-Token-based Patch Selection Method}
\label{sec:auto_patch_sel}
The limitation of the first two patch selection methods is that they rely on the performance of the lightweight off-the-shelf pose estimator. This limitation can become especially problematic when dealing with complex scenes, as the accuracy of the pose estimator is often compromised by occlusion, motion, or variations in camera perspective. As a consequence, this might result in the selection of irrelevant patches and removing important patches, leading to suboptimal performance.
Therefore, a more robust approach is required that can adapt to these challenging scenes without the need for an off-the-shelf pose estimator. 

\myparagraph{Selecting most informative patches via learnable joint tokens}
We propose to overcome this limitation by selecting patches using the learnable joint tokens,
which serve as a powerful feature representation for distinguishing the relevant body part patches.
Specifically, we aim to determine the importance of each patch in relation to the joint tokens, thereby enabling us to select the most informative body part patches. To achieve this, we harness the attention matrix similar to~\cite{Liang2022NotAP, Fayyaz2022AdaptiveTS, Goyal2020PoWERBERTAB}, as the values in $\mathcal{A}$ serve as the weight of contribution of input tokens to output tokens. For example, $\mathcal{A}_{N+1,1:N}$ denotes the contribution weights of the $N$ patch tokens to the $(N+1)^{\text{th}}$ output token, \ie first joint token. Thus, we can calculate the average contribution weight of a patch token $l$ to the $J$ joint tokens, as follows:
\begin{equation}
\mathcal{W}_l = \frac{1}{J} \sum_{j=1}^J \mathcal{A}_{N+j, l}.
\end{equation}
Following~\cite{Fayyaz2022AdaptiveTS}, we take the norm of $\mathcal{V}_l$ into account for calculating the importance score. Thus, the importance score of the patch token $l$ is:
\begin{equation}
\mathcal{I}_l = \frac{\mathcal{W}_l\times {\|\mathcal{V}_l\|}}{ \sum_{k=1}^N \mathcal{W}_k \times \|\mathcal{V}_k\|},
\end{equation}
where $l,k \in \{1,\dots,N\}$.
Once the importance scores of each patch token have been computed, we select $L$ patch tokens with the highest scores for further processing, where $L \ll\ N$. 

\myparagraph{Pruning attention matrix} Our subsequent step involves pruning the attention matrix $\mathcal{A}\in \mathbb{R}^{(N+J)\times (N+J)}$ by selecting the rows that correspond to the chosen $L$ patch tokens and $J$ joint tokens, designated as $\mathcal{A}^s \in \mathbb{R}^{(L+J)\times (L+J)}$. We then compute the output tokens $\mathbf{O}^s\in \mathcal{R}^{(L+J)\times C}$, given by:
\begin{equation}
\mathbf{O}^s = \mathcal{A}^s \mathcal{V}^s.
\end{equation}
These output tokens are then passed as input for the next blocks. 

\myparagraph{Refining background patches via joint tokens} Although only $\mathbf{O}^s$ will be processed in the next blocks of ViT, the non-selected patch tokens will still be used during the heatmap decoding.
Therefore, it is important to have a refined representation of the non-selected patch tokens before they are excluded from further processing in the next blocks. Thus, we propose an efficient method that updates these tokens using the joint tokens. This approach is motivated by the fact that joint tokens learn global information and therefore can be used to update the patch tokens in a computationally efficient manner without the need to compute contributions from all tokens, which can be computationally expensive. We start by selecting the rows of the attention matrix $\mathcal{A}$ that correspond to the non-selected patch tokens and the columns that correspond to the joint tokens, resulting in a sub-matrix $\mathcal{A}^o \in \mathbb{R}^{(N-L) \times J}$. We then update the non-selected patch tokens using the joint tokens as follows:
\begin{equation}
\mathbf{O}^o = \mathcal{A}^o \mathcal{V}^j,
\end{equation}
where $ \mathcal{V}^j$ corresponds to the values of the joint tokens. 

\section{Experiments}
\label{sec:experiment}
\subsection{Implementation details}

In our experiments, we employ the common top-down setting for human pose estimation.
We follow most of the default training and evaluation settings of the mmpose~\cite{mmpose2020}
framework but use the AdamW optimizer with a learning rate of $5e-4$ and UDP as a post-processing method.
We use ViT-B and ViT-L as backbones and refer to the corresponding models as \method-B and  \method-L. The backbones are pre-trained with MAE~\cite{He2022MaskedAA} weights and trained with multiple datasets as in~\cite{Xu2022ViTPoseSV}. 

\subsection{Dataset details}
The proposed methods are evaluated on six 2D pose estimation benchmarks: namely MPII~\cite{Andriluka20142DHP}, COCO~\cite{Lin2014MicrosoftCC}, AI Challenger~\cite{Wu2017AIC}, CrowdPose~\cite{Li2018CrowdPoseEC}, JRDB-Pose~\cite{Vendrow2022JRDBPoseAL}, and OCHuman~\cite{Zhang2019Pose2SegDF}. The datasets are challenging and diverse, with varying numbers of images, person instances, and annotated joints. The first five datasets are used to train and test the proposed method, and OCHuman is only used to test the models in dealing with occluded scenes.

\subsection{Evaluation metrics}
On the MPII benchmark, we adopt the standard PCKh metric as our performance evaluation metric. PCKh~\cite{Andriluka20142DHP} is an accuracy metric that measures if the predicted joint and the true joint are within a certain distance threshold ($50\%$ of the head segment length). On the remaining benchmarks, we adopt standard average precision (AP) as our main performance evaluation metric. AP is calculated using Object Keypoint Similarity (OKS), which measures how close the predicted joint location is to the ground-truth joint. Additionally, we will evaluate the computational complexity of the models by measuring the total number of Floating Point Operations (FLOPs) that each model needs to perform.

\begin{table}[tb]
    \centering    
    \caption{Performance of the proposed patch selection methods on three benchmarks, namely COCO val set, MPII test set, and OCHuman test set. The best results among the base and large model variants are highlighted in bold.\\}
    \label{tab:multi_v1}
  \begin{tabular}{@{}llccccc@{}}
    \toprule
    Model & Patch Selection &  Params & FLOPs & COCO & MPII & OCH \\
    \midrule    
    Lite-HRNet~\cite{Yu2021LiteHRNetAL} & None & 1M & ~~0.2G & 64.8 & 86.1 & 51.9\\
    SimpleBaseline~\cite{Xiao2018SimpleBF} & None & 69M & 15.7G & 72.0 & 89.0 & 58.2 \\
    HRNet-W48~\cite{Sun2019DeepHR} & None & 64M & 14.6G & 75.1 & 90.1 & 60.4 \\
    TransPose-H/A6~\cite{Yang2021TransPoseKL} & None & 18M & 21.8G & 75.8 & 92.3 & - \\
    TokenPose-L/D24~\cite{Li2021TokenPoseLK} & None & 28M & 11.0G & 75.8 & - & - \\
    HRFormer-B~\cite{Yuan2021HRFormerHT} & None & 43M & 12.2G & 75.6 & - & 49.7 \\
    ViTPose-B~\cite{Xu2022ViTPoseSV} & None & 86M & 18.0G & 77.1 & \textbf{93.3} & 87.3 \\
    ViTPose-L~\cite{Xu2022ViTPoseSV} & None & 307M & 59.8G & 78.7 & \textbf{94.0} & 90.9 \\
    ViTPose-H~\cite{Xu2022ViTPoseSV} & None & 632M & 122.8G & 79.5 & 94.1 & 90.9 \\
    \midrule 
    \method-B (Ours) & None &  90M & 19.8G & \textbf{77.6} & 92.4 & \textbf{93.0} \\
    \method-B (Ours) & Neighbors &  90M & 11.1G & 74.1 & 91.8 & 89.5 \\
    \method-B (Ours) & Skeleton &  90M & 13.3G & 75.0  & 92.1 & 90.1 \\
    \method-B (Ours) & Joint Tokens &  90M & 13.7G & \textit{76.5} & {92.5} & \textit{92.3}  \\
    \midrule
    \method-L (Ours) & None &  309M & 66.7G & \textbf{78.7} & {92.8} & \textbf{94.3} \\
    \method-L (Ours) & Neighbors &  309M & 35.6G & 75.7 & 92.1 & 90.0 \\
    \method-L (Ours)& Skeleton &  309M & 38.3G & 76.3 & 92.4 & 92.6 \\
    \method-L (Ours) & Joint Tokens &  309M & 45.5G &  \textit{77.3} &  \textit{92.7} & \textit{93.6}  \\
    \bottomrule
  \end{tabular}
\end{table}

\begin{table}[tbh]
\centering
    \caption{Performance of the proposed joint-token-based patch selection method on the other three benchmarks, namely AI Challenger val set, CrowdPose test set, and JRDB-Pose val set.\\}
    \label{tab:multi_v2}
  \begin{tabular}{@{}llccccc@{}}
    \toprule
    Model & Patch Selection &  Params & FLOPs & AIC & CPose & JRDP \\
    \midrule    
    SimpleBaseline~\cite{Xiao2018SimpleBF} & None & 69M & 15.7G & 29.9  & 60.8  & - \\
    HRNet-W48~\cite{Sun2019DeepHR} & None & 64M & 14.6G & 33.5 & - & 42.4 \\
    HRFormer-B~\cite{Yuan2021HRFormerHT} & None & 43M & 12.2G & 34.4 & 72.4 & -  \\
    \midrule 
    \method-B (Ours) & None &  90M & 19.8G & \textbf{36.6} & \textbf{76.3} & \textbf{73.9}  \\
    \method-B (Ours) & Joint Tokens &  90M & 13.7G & \textit{35.0} & \textit{74.5} & \textit{72.8}  \\
    \midrule
    \method-L (Ours) & None &  309M & 66.7G & \textbf{38.3}  & \textbf{77.9}  & \textbf{74.9}  \\
    \method-L (Ours) & Joint Tokens &  309M & 45.5G & \textit{37.1} & \textit{76.7} & \textit{74.2 }\\
    \bottomrule
  \end{tabular}
\end{table}

\subsection{Results}
\label{sec:results}
\begin{figure}[tb]
    \includegraphics[width=0.9\columnwidth,clip=true, trim = 0 29 0 12]{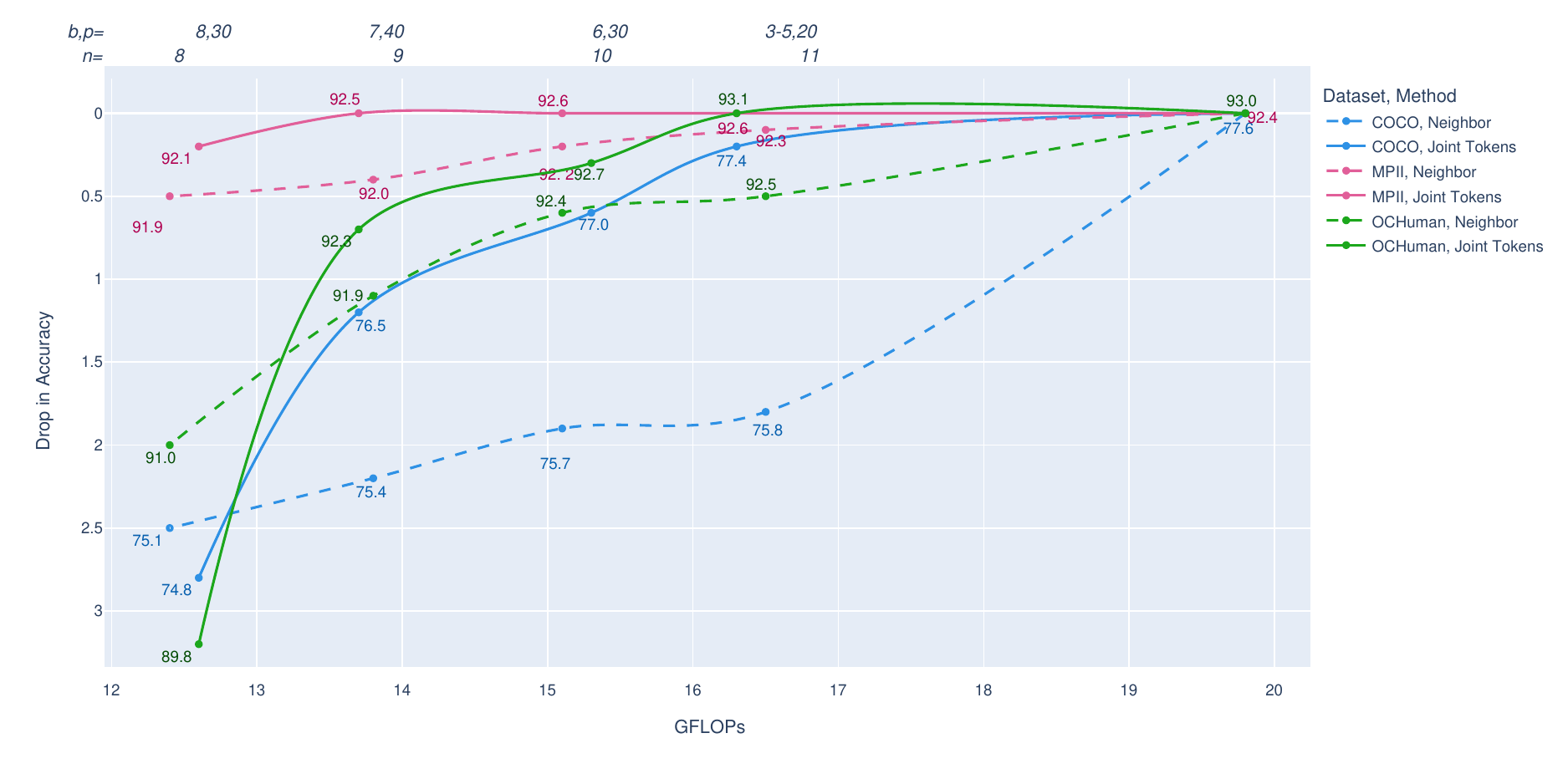}
  \caption{\textbf{Trade-off between accuracy and GFLOPs on three benchmarks: COCO, MPII, and OCHuman --} 
  The performance of \method-B with two patch selection methods: Neighbors (dashed line) and Joint-Token-based (solid line).
  $n$ denotes to the number of neighbors selected and $b,p$ refers to $p$ number of patches that are removed at block $b$ in the Joint Tokens method.}
  \label{fig:tradeoff}
 
\end{figure}

\begin{figure}[ht]
\centering
\includegraphics[width=0.99\columnwidth]{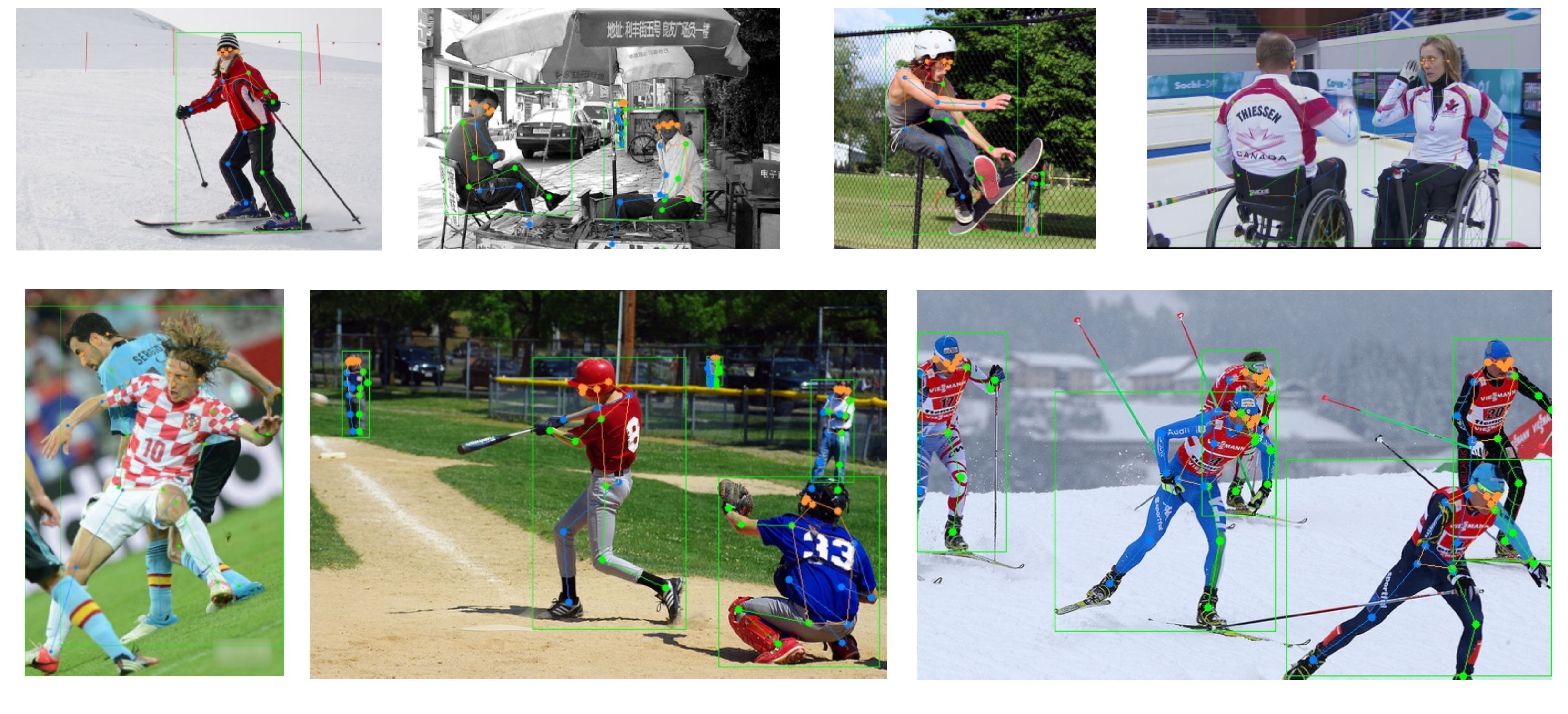}
\caption{Qualitative results on sample images from the evaluation benchmarks.}
\label{fig:qual_results}
\end{figure}

The performance of our proposed methods and other convolutional and transformer-based methods on three datasets, namely the COCO val set, the MPII test set, and the OCHuman test set, are presented in Table \ref{tab:multi_v1}. We used LiteHRNet~\cite{Yu2021LiteHRNetAL}, a lightweight and less accurate pose estimation network, to guide the first two patch selection methods. Additional results of the joint-token-based patch selection method on AI Challenger val set, CrowdPose test set, and JRDB-Pose val set is presented in Table \ref{tab:multi_v2}. The remaining patch selection methods were not evaluated on these benchmarks as we were not able to find lightweight pose estimators trained on them.  
Our experiments show that \method\ without patch selection outperforms all, but is computationally expensive. However, our proposed patch selection methods proves to be beneficial in this regard, as it significantly reduces computational costs while maintaining high accuracy. For instance, we achieve a reduction of 30\% to 44\% in GFLOPs with a slight drop in accuracy ranging from 1.1\% to 3.5\% for COCO, 0\% to 0.6\% for MPII, and 0.7\% to 3.5\% for OCHuman. 
We can also control the drop in accuracy by changing the number of patches to be selected. The trade-off between performance and computational complexity for the neighboring and joint-token-based patch selection methods is depicted in Figure~\ref{fig:tradeoff}. The neighboring and skeleton patch selection remove irrelevant patches before they are processed by ViT, while the joint-token-based selection method learns to remove them on the fly. Thus, the first two approaches prioritize efficiency over accuracy by removing patches early on. For example, they are effective for addressing the low end range in Figure~\ref{fig:tradeoff}, where the joint-token-based selection method performs poorly. 
Some qualitative results of our method 
on sample images from the benchmarks are illustrated in Figure \ref{fig:qual_results}.

\begin{figure}[tb]
    \includegraphics[width=0.99\columnwidth,clip=true, trim = 0 30 0 70]{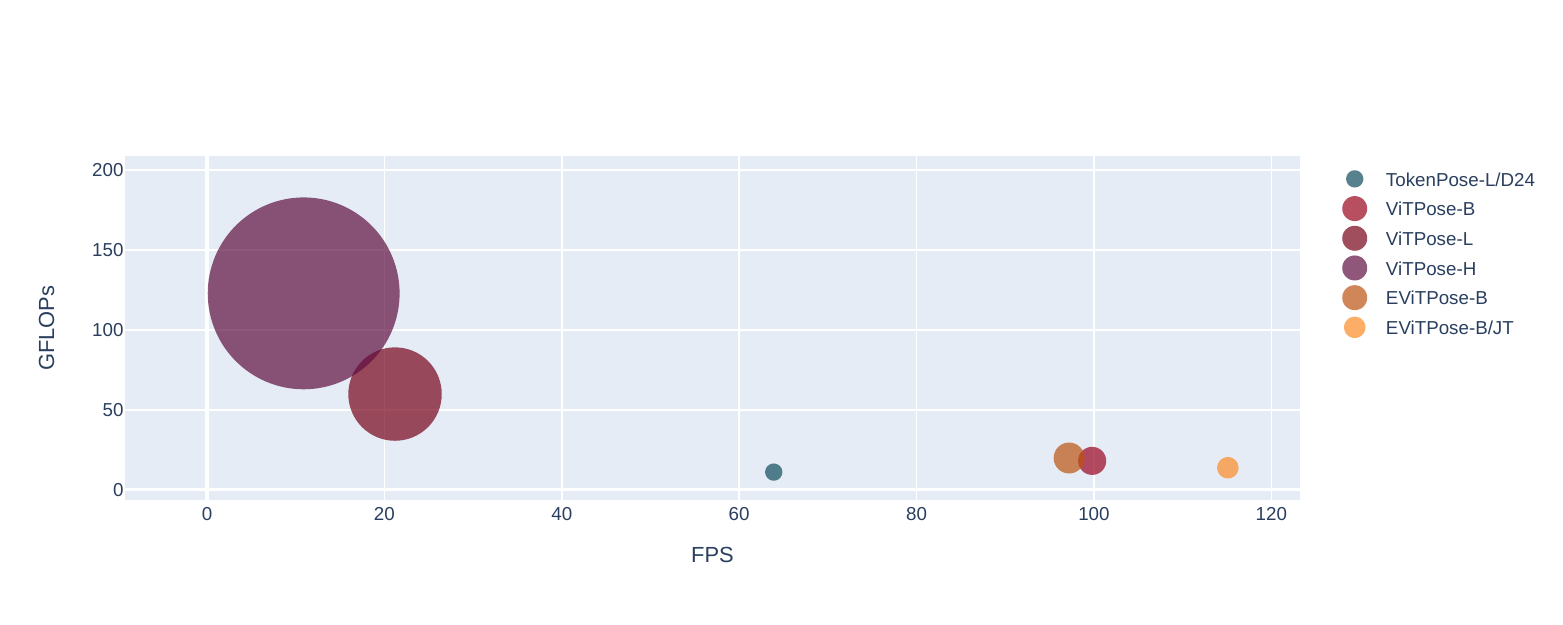}
  \caption{\textbf{Runtime (FPS) vs GFLOPs comparison --}
  The Joint-Token-based Patch Selection method (\method-B/JT) achieves an 88\% reduction in GFLOPs and a 10$\times$ (955\%) increase in FPS compared to ViTPose-H, with a minimal accuracy drop of up to 2.9\%.
  }
  \label{fig:runtime_analysis}
 
\end{figure}

\myparagraph{Ablation} We conducted a run-time comparison (measured in frames per second, FPS) among \method, ViTPose and TokenPose, presented in Table~\ref{fig:runtime_analysis}. The results show that our Joint-Token-based Patch Selection method (\method-B/JT) achieves an 88\% reduction in GFLOPs and 10$\times$ 
increase in FPS with respect to ViTPose-H, with a minimal drop in accuracy of upto 2.9\%.
Furthermore, we evaluate the effect of the lightweight pose network in the performance of \method\ with the first two patch selection methods as shown in Table~\ref{tab:lw_ablation}. \method\ maintains consistent performance across three different lightweight pose estimation models, namely LiteHRNet~\cite{Yu2021LiteHRNetAL}, MobileNetv2~\cite{toshev2014deeppose, sandler2018mobilenetv2}, and ShuffleNet~\cite{toshev2014deeppose, zhang2018shufflenet}.  

\begin{table}[hbt]
  \centering
  \caption{Effect of lightweight pose network in performance of \method\ on the MPII dataset.\\}
  \label{tab:lw_ablation}
  \begin{tabular}{@{}ccc|ccc@{}}
    \toprule
    \multicolumn{3}{c}{Neighbors} & \multicolumn{3}{c}{Skeleton}  \\
    LiteHRNet & MobileNetv2 & ShuffleNet & LiteHRNet & MobileNetv2 & ShuffleNet \\
    \midrule
    92.1 & 91.9 & 92.0 & 91.8 & 91.8 & 91.8 \\
    \bottomrule
  \end{tabular}
\end{table}

\section{Conclusion}
\label{sec:conclusion}

In this work, we have proposed \method, a Vision Transformer-based human pose estimation network with patch selection methods that greatly reduces the computational complexity of ViTs while maintaining high accuracy.
The proposed methods involve selecting and processing a small subset of patches that contain the most important information about body joints. Two of the methods utilize a fast yet imprecise out-of-the-shelf pose estimation network to guide the patch selection process, while the third method uses learnable joint tokens to progressively select the most informative patches. 
The experimental results on six widely-used 2D pose estimation benchmarks demonstrate that our methods significantly improve speed and reduce computational complexity, with reductions ranging from 30\% to 44\% in GFLOPs, with only a slight drop in accuracy, between 0\% and 3.5\%.

\myparagraph{Acknowledgments} The authors thank Carolina Pacheco and Yutao Tang for their
valuable input and feedback throughout the development of
this work. This research is based upon work supported in part by the Office of the Director of National Intelligence (ODNI), Intelligence Advanced Research Projects Activity (IARPA), via [2022-21102100005]. The views and conclusions contained herein are those of the authors and should not be interpreted as necessarily representing the official policies, either expressed or implied, of ODNI, IARPA, or the U.S. Government. The U.S. Government is authorized to reproduce and distribute reprints for governmental purposes notwithstanding any copyright annotation therein.

\bibliography{egbib}
\end{document}